\PassOptionsToPackage{most}{tcolorbox}
\PassOptionsToPackage{table}{xcolor}
\documentclass[11pt]{article}

\usepackage[final]{arxiv}

\usepackage{times}
\usepackage{latexsym}

\usepackage[utf8]{inputenc}
\usepackage{amsmath,amssymb}
\usepackage{pifont}
\usepackage{placeins} 
\usepackage{booktabs}
\usepackage{multirow}
\usepackage{lipsum}
\usepackage{tabularx}
\usepackage{subcaption}
\usepackage{arydshln}
\usepackage{adjustbox} 
\usepackage{makecell}

\usepackage{siunitx}
\usepackage[T1]{fontenc}

\usepackage{inconsolata}
\usepackage{dblfloatfix}
\usepackage{graphicx}
\usepackage{enumitem}
\usepackage{cuted}
\usepackage{capt-of} 
\usepackage{graphicx}
\usepackage{dblfloatfix} 
\usepackage{caption}  
\usepackage{fontawesome5} 
\usepackage{hyperref}     

\usepackage{graphicx}
\usepackage{tikz}
\usetikzlibrary{fadings} 
\usepackage{wrapfig}
\usepackage{svg}
\usepackage{booktabs}
\usepackage{multirow}
\usepackage{siunitx}

\usepackage[utf8]{inputenc}
\usepackage[table]{xcolor} 
\usepackage{array}
\usepackage[most]{tcolorbox}
\usepackage{tcolorbox}
\usepackage{tabularx}
\usepackage{adjustbox}
\usepackage{makecell}
\usepackage{CJKutf8}
\usepackage{svg}
\usepackage{threeparttable}

\definecolor{navyblue}{HTML}{0071BC}

\newtcolorbox{promptbox}[1][]{
    colback=gray!5!white,      
    colframe=gray!75!black,    
    title=\textbf{Prompt Template}, 
    fonttitle=\bfseries\sffamily,
    fontupper=\ttfamily\small, 
    sharp corners, rounded corners=southeast, arc=3mm, 
    boxrule=1pt,               
    enhanced,                  
    breakable,                 
    #1                         
}

\newtcolorbox{keyfinding}[1][]{
  colback=cyan!5!white,    
  colframe=cyan!75!black,  
  fonttitle=\bfseries,     
  title={Key Findings},    
  #1                       
}

\makeatletter
\AtBeginDocument{%
  \renewcommand{\@cite}[2]{\textsuperscript{[#1]}}%
}
\makeatother

\title{\textsc{3D-Mix} for VLA: A Plug-and-Play Module for Integrating VGGT-based 3D Information into Vision-Language-Action Models}

\author{
  Bin Yu\textsuperscript{1,2,\thanks{Equal contribution}}
  Shijie Lian\textsuperscript{2,4,\footnotemark[1]}
  Xiaopeng Lin\textsuperscript{2,5,\footnotemark[1]}
  Zhaolong Shen\textsuperscript{2,6,\footnotemark[1]}
  Yuliang Wei\textsuperscript{1,\thanks{Corresponding author}}
  Haishan Liu\textsuperscript{2}\\
  \textbf{Changti Wu\textsuperscript{2,7}}
  \textbf{Hang Yuan\textsuperscript{2,7}}
  \textbf{Bailing Wang\textsuperscript{1}}
  \textbf{Cong Huang\textsuperscript{2,3}}
  \textbf{Kai Chen\textsuperscript{2,3,8,\footnotemark[2]}}
  \\[2ex]
  \textsuperscript{1}HIT\quad
  \textsuperscript{2}ZGCA\quad
  \textsuperscript{3}ZGCI\quad
  \textsuperscript{4}HUST\quad
  \textsuperscript{5}HKUST(GZ)\quad
  \textsuperscript{6}BUAA\quad
  \textsuperscript{7}ECNU\quad
  \textsuperscript{8}DeepCybo
}

\begin{document}
\maketitle

\begin{center}
    \vspace{-50pt}
    \faGithub\hspace{6pt}\href{https://github.com/ZGC-EmbodyAI/3DMix-for-VLA}{\texttt{\color{black}https://github.com/ZGC-EmbodyAI/3DMix-for-VLA}}
\end{center}

\vspace{5pt}

\newcommand{\methodname}{\textsc{3D-Mix}}
\newcommand{\gain}[1]{\,\textcolor{green!60!black}{$\uparrow$\scriptsize{#1}}}
\newcommand{\gainph}[1]{\phantom{\,\textcolor{green!60!black}{$\uparrow$\scriptsize{#1}}}}

\begin{abstract}
Vision-Language-Action (VLA) models leverage Multimodal Large Language Models (MLLMs) for robotic control, but recent studies reveal that MLLMs exhibit limited spatial intelligence due to training predominantly on 2D data, resulting in inadequate 3D perception for manipulation tasks. While recent approaches incorporate specialized 3D vision models such as VGGT to enhance spatial understanding, they employ diverse integration mechanisms without systematic investigation, leaving the optimal fusion strategy unclear. We conduct a comprehensive pilot study comparing nine VGGT integration schemes on standardized benchmarks and find that \textit{semantic-conditioned gated fusion}, which adaptively balances 2D semantic and 3D geometric features based on task context, achieved the strongest performance among all nine evaluated fusion schemes in our pilot study. We present \textbf{\methodname{}}, a plug-and-play module that integrates into diverse VLA architectures (GR00T-style and $\pi$-style) without modifying existing MLLM or action expert components. Experiments across six MLLM series (nine model variants, 2B--8B parameters) on SIMPLER and LIBERO show that \methodname{} delivers consistent performance gains, averaging $+7.0\%$ on the out-of-domain (OOD) SIMPLER benchmark across all nine GR00T-style variants, establishing a principled approach for enhancing spatial intelligence in VLA systems.
\end{abstract}

\section{Introduction}
\label{sec:intro}

Vision-Language-Action (VLA) models have emerged as a promising paradigm for robotic manipulation, unifying visual perception, language understanding, and action generation within end-to-end learned systems. By leveraging Multimodal Large Language Models (MLLMs) as semantic encoders coupled with specialized action prediction modules, VLA models enable robots to follow natural language instructions and generalize across diverse tasks.

Despite their success, current VLA models face a fundamental limitation: MLLMs are predominantly pre-trained on 2D image-text corpora lacking explicit 3D geometric supervision. As a result, they exhibit limited depth perception and struggle with precise spatial reasoning, capabilities critical for manipulation tasks that require accurate localization, grasp pose estimation, and spatial relationship understanding.

To address this limitation, recent research has explored integrating specialized 3D vision models into VLA architectures. Among various 3D encoders, VGGT (Visual Geometry Grounded Transformer)~\citep{VGGT_2025} has emerged as a compelling choice due to its ability to extract rich geometric features from single or multi-view images in a unified framework. However, existing approaches to incorporating VGGT into VLA systems employ diverse fusion strategies without systematic comparison, leaving the optimal integration strategy unclear. This raises several questions: \textit{Where} should 3D features be injected? \textit{How} should geometric and semantic information be combined? \textit{What} fusion mechanisms enable effective cross-modal interaction?

We address these questions through a comprehensive empirical study of VGGT integration strategies. We systematically evaluate nine distinct fusion schemes spanning the entire VLA pipeline and identify the key factors that determine fusion effectiveness. We then propose \textbf{\methodname{}}, a plug-and-play module that integrates VGGT-derived 3D features through semantic-conditioned adaptive gating. Unlike fixed fusion strategies, \methodname{} employs learnable gates that dynamically balance semantic and geometric information at each spatial location, enabling context-aware fusion without modifying MLLM or action expert internals.

Experiments across six MLLM series (nine model variants, 2B--8B parameters; Qwen2.5-VL, Qwen3-VL, RoboBrain2/2.5, MimoEmbodied, RynnBrain) on SimplerEnv and LIBERO show that \methodname{} consistently improves performance across architectures and model scales.

Our contributions are as follows:
\begin{itemize}
    \item We present the first systematic study of VGGT integration schemes for VLA models, evaluating nine fusion strategies to understand how 3D information contributes to manipulation performance.
    \item We propose \textbf{\methodname{}}, a lightweight plug-and-play module that performs principled 3D-semantic fusion through adaptive gating, achieving consistent improvements without architecture-specific modifications.
    \item We demonstrate \methodname{}'s generalizability across multiple VLA architectures and MLLM backbones, providing practical guidelines for enhancing spatial intelligence in VLA systems.
\end{itemize}

\section{Related Work}
\label{sec:related}

\paragraph{Vision-Language-Action Models.}
Vision-Language-Action (VLA) models unify perception, reasoning, and control within end-to-end frameworks by leveraging pre-trained MLLMs for generalizable robot policies. Recent works such as OpenVLA~\citep{OpenVLA_24}, $\pi_{0.5}$~\citep{PI0, PI05_25}, and GR00T-N1.6~\citep{GR00T_N1.6} have demonstrated effective robotic control by scaling MLLMs with robotic data under the VLA architecture. The VLA architectures used in this paper build upon these works.

\paragraph{3D Fusion for VLA Models.}
While VLAs are built upon pretrained MLLMs, recent works~\citep{SpatialBlindSpot_26,SpatialTree_26,CambrianS_25} have revealed the limited spatial reasoning capabilities of MLLMs, attributing this primarily to the inability to explicitly model 3D structure. VGGT~\citep{VGGT_2025} has attracted attention as a pretrained Transformer applicable to diverse 3D downstream tasks. Several works~\citep{SpatialForcing_25,Evo0_25,ABot-M0_26} have accordingly introduced 3D information from VGGT into VLA systems to improve robotic control. However, these works adopt diverse fusion strategies without systematic comparison, leaving the community without principled guidance on how to best incorporate 3D information. We address this gap with a comprehensive pilot study of nine fusion schemes and distill the empirical insights into \textbf{\methodname{}}, a lightweight plug-and-play module.


\section{Pilot Study: Comparative Analysis of VGGT Fusion Schemes}
\label{sec:pilot_study}

To systematically investigate the optimal approach for integrating VGGT-derived 3D geometric information into VLA models, we compare nine distinct fusion schemes in a controlled pilot study. This study identifies the most effective architectural pattern before scaling to full experiments. We first describe the base VLA architecture, then detail each fusion scheme, and finally present the experimental setup.

\begin{figure}[h]
    \centering
    \includegraphics[width=1.0\textwidth]{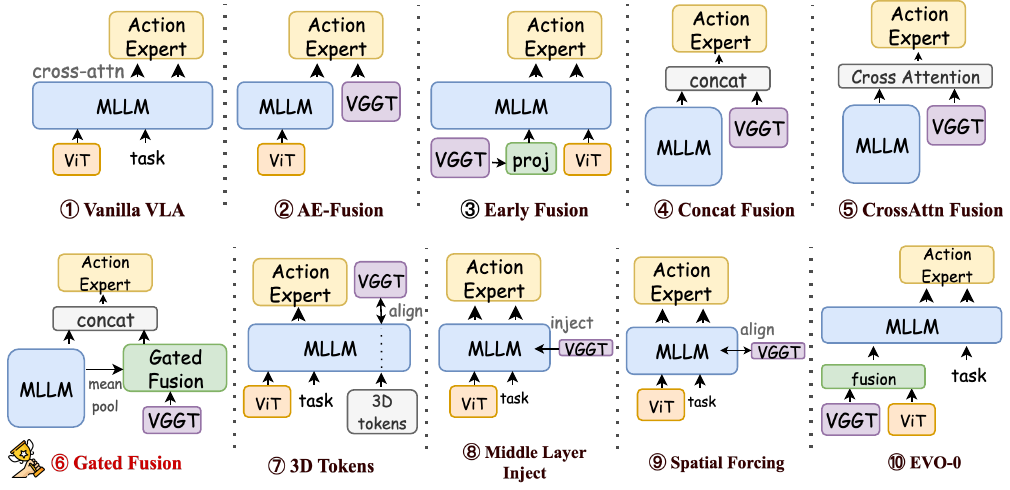}
    \caption{\textbf{Overview of nine VGGT fusion schemes} evaluated in our pilot study. Each scheme integrates VGGT-derived 3D geometric features into the VLA pipeline at a different location or via a different mechanism, ranging from early token-level injection to training-time alignment and cross-attention fusion.}
    \label{fig:all-fusions}
\end{figure}

\subsection{Base VLA Architecture}
\label{sec:base_vla}

We adopt a GR00T-style VLA framework comprising an MLLM backbone (Qwen3-VL~\citep{Qwen3-VL}) that encodes multi-view images and language instructions into hidden states $\mathbf{H} \in \mathbb{R}^{B \times L \times D}$, and a DiT-based flow-matching action expert~\citep{DiT_23} that predicts future actions through a flow-matching objective. This modular architecture enables systematic evaluation of different 3D fusion mechanisms without disrupting other components. Detailed architecture specifications are provided in Appendix~\ref{sec:appendix_groot}.

\subsection{Multiple VGGT Fusion Schemes}
\label{subsec:fusion_schemes}

We investigate nine distinct fusion schemes for integrating VGGT-derived 3D geometric features into the base VLA architecture, as illustrated in Figure~\ref{fig:all-fusions}. Each scheme differs in \emph{where} and \emph{how} VGGT features are injected, spanning from early input-level fusion to late action-stage fusion:

\textbf{(1) AE-Fusion}: Augments the action expert with dual cross-attention to simultaneously attend to MLLM semantic features and VGGT geometric features during denoising.

\textbf{(2) Early Fusion}: Injects VGGT tokens directly into the MLLM input sequence, enabling implicit semantic-geometric interaction via self-attention.

\textbf{(3) Concat Fusion \& (4) CrossAttn Fusion}: Apply GateMixer preprocessing to VGGT features, then either concatenate with MLLM outputs (Concat) or apply explicit cross-attention before concatenation (CrossAttn).

\textbf{(5) Gated Fusion}: Employs learnable gates to dynamically balance semantic and geometric features at each token position based on global semantic context.

\textbf{(6) 3D-Tokens}: Appends a special token $\langle|\text{vggt}|\rangle$ supervised to encode 3D geometry via alignment loss; VGGT not required at inference.

\textbf{(7) Middle Layer Injection}: Injects VGGT features into an intermediate MLLM layer via adapter-style cross-attention, enabling deeper semantic-geometric integration.

\textbf{(8) Spatial Forcing}~\citep{SpatialForcing_25}: Supervises intermediate MLLM layers with alignment loss during training; discards VGGT at inference with zero overhead.

\textbf{(9) Visual Fusion}: Fuses VGGT 3D tokens with MLLM 2D visual tokens via cross-attention before passing to the MLLM backbone.  

Detailed technical descriptions of each fusion scheme are provided in Appendix~\ref{sec:appendix_fusion_details}.

\subsection{Experimental Setup for Comparative Evaluation}
\label{subsec:pilot_setup}

To fairly compare the nine fusion schemes, we establish a standardized experimental protocol with consistent model architectures, training procedures, and evaluation benchmarks.

We use Qwen3-VL-4B~\citep{Qwen3-VL} as the vision-language backbone and VGGT-1B~\citep{VGGT_2025} as the frozen 3D feature extractor. The VGGT encoder takes multi-view RGB images as input and produces a set of geometric tokens that are subsequently consumed by each fusion scheme; its weights are kept frozen throughout training to isolate the effect of fusion design from 3D encoder quality.

\textbf{Training Configuration.}
All models are trained for 60{,}000 steps on 8$\times$ NVIDIA H100 GPUs with DeepSpeed ZeRO-2~\cite{deepspeed_2020}. We use a per-device batch size of 16 (effective global batch size of 128). The MLLM backbone is optimized with a learning rate of $10^{-5}$, while the action expert and any fusion-specific parameters are trained with a learning rate of $10^{-4}$; both use the AdamW optimizer~\cite{adamw_2017} and a cosine learning rate schedule with linear warm-up.

\textbf{Datasets and Benchmarks.}
We evaluate each fusion scheme on two complementary settings. In the first, models are trained on the BridgeV2 subset of Open X-Embodiment~\cite{OXE_24} and evaluated on SimplerEnv~\cite{SimplerEnv_24} (hereafter SIMPLER), a real-to-simulation (real2sim) OOD benchmark measuring cross-domain generalization to unseen object configurations and visual variations. In the second, models are trained and evaluated on LIBERO~\cite{libero} official expert trajectories across four task suites (Spatial, Object, Goal, Long-Horizon), assessing in-domain multi-task learning. The primary metric is average task success rate within each benchmark.

\begin{table}[htbp]
    \centering
    \caption{\textbf{Comparison of nine VGGT fusion schemes} on the SIMPLER (out-of-domain, real-to-sim) and LIBERO (in-domain, multi-task) benchmarks.}
    \label{tab:results}
    \small
    \begin{tabular}{lcccccc}
        \toprule
        \multirow{2}{*}{\textbf{Method}} & \multicolumn{4}{c}{\textbf{SIMPLER Tasks}} & \multirow{2}{*}{\textbf{SIMPLER Avg}} & \multirow{2}{*}{\textbf{LIBERO Avg}} \\
        \cmidrule(lr){2-5}
        & Stack Green & Put Carrot & Put Spoon & Put Eggplant & & \\
        \midrule
        \rowcolor{gray!15} Base (Qwen3-VL-4B-Instruct) & 22.92 & 57.29 & 62.50 & 88.54 & 57.81 & 96.5 \\
        + AE Fusion     & 0.00 & 0.00 & 0.00 & 12.50 & \textcolor{red}{3.13} & \textcolor{green!60!black}{97.40} \\
        + 3D-Tokens      & 26.04 & 53.12 & 76.04 & 67.79 & \textcolor{red}{56.25} & \textcolor{green!60!black}{97.64} \\
        + Early Fusion   & 5.21 & 40.62 & 48.96 & 83.33 & \textcolor{red}{44.53} & \textcolor{red}{86.45} \\
        + Visual Fusion  & 0.00 & 3.12 & 1.04 & 14.58 & \textcolor{red}{4.69} & \textcolor{red}{73.40} \\
        \rowcolor{navyblue!10} \textbf{+ GatedFusion} \includegraphics[height=1em]{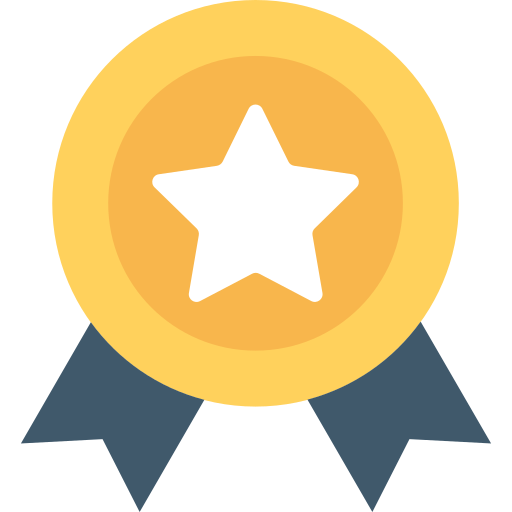} & 30.21 & 61.46 & 86.46 & 94.79 & \textbf{\textcolor{green!60!black}{68.23}} & \textbf{\textcolor{green!60!black}{98.05}} \\
        + Concat Fusion & 16.67 & 52.08 & 75.00 & 97.92 & \textcolor{green!60!black}{60.42} & \textcolor{green!60!black}{97.75} \\
        + CrossAttn Fusion & 31.94 & 43.06 & 70.83 & 79.17 & \textcolor{red}{56.25} & \textcolor{green!60!black}{83.45} \\
        + Middle Layer Injection & 11.46 & 48.96 & 70.83 & 76.04 & \textcolor{red}{51.82} & \textcolor{green!60!black}{97.82} \\
        + Spatial Forcing & 25.00 & 52.08 & 73.96 & 84.37 & \textcolor{green!60!black}{58.85} & \textcolor{green!60!black}{97.72} \\
        \bottomrule
    \end{tabular}
\end{table}

\subsection{Pilot Study Results and Analysis}
\label{subsec:pilot_results}

Table~\ref{tab:results} presents the evaluation results of nine VGGT fusion schemes on SIMPLER and LIBERO.

On \textbf{SIMPLER (Real-to-Sim OOD)}~\citep{SimplerEnv_24}, which measures cross-domain generalization, GatedFusion achieves the highest average success rate of 68.23\%, a 10.42\% gain over the base model (57.81\%). This confirms that the semantic-conditioned gating mechanism selectively incorporates 3D geometric cues for spatial reasoning.

On \textbf{LIBERO (In-Domain)}~\citep{libero}, which evaluates in-domain multi-task learning, GatedFusion again achieves the highest performance (98.05\%) among the evaluated schemes. Its consistent advantage across both benchmarks highlights the robustness of this fusion strategy.

We therefore adopt \textbf{GatedFusion} as the foundation for \textbf{\methodname{}}. We implement \methodname{} as a plug-and-play component that integrates into diverse VLA architectures. In the following sections, we validate its generalizability across both GR00T-style and $\pi$-style architectures and multiple MLLM backbones.

\section{\methodname{} Module}
\label{sec:method}

In this section, we present the \textbf{\methodname{}} module, a plug-and-play component that adapts the GatedFusion strategy to mainstream VLA architectures.

\subsection{Module Design}

The \methodname{} module operates on the principle of \textit{semantic-conditioned adaptive gating}, enabling dynamic fusion of 2D semantic features from VLMs with 3D geometric features from VGGT. Different manipulation tasks require varying degrees of reliance on semantic understanding versus geometric precision. Rather than using a fixed fusion ratio, \methodname{} uses a learnable gating mechanism that adaptively determines the optimal balance at each token position.

\textbf{Feature Extraction and Projection.}
Given multi-view RGB images $\mathcal{I} = \{I_1, \dots, I_V\}$, VGGT extracts geometry-aware patch tokens $\mathbf{F}_{\text{VGGT}} \in \mathbb{R}^{B \times N_{\text{patches}} \times D_{\text{VGGT}}}$. To align these geometric features with the MLLM's representation space, we apply a linear projection:
\begin{equation}
\mathbf{F}_{\text{geo}} = \mathbf{W}_{\text{proj}} \mathbf{F}_{\text{VGGT}}, \quad \mathbf{F}_{\text{geo}} \in \mathbb{R}^{B \times N_{\text{patches}} \times D},
\end{equation}
where $D$ is the MLLM hidden dimension. Simultaneously, the MLLM processes the same images along with language instruction $\ell$ to produce semantic hidden states $\mathbf{H}_{\text{MLLM}} \in \mathbb{R}^{B \times L \times D}$.

\textbf{Semantic-Conditioned Gating.}
To enable context-aware fusion, we first extract a global semantic summary by mean-pooling the MLLM hidden states:
\begin{equation}
\mathbf{s}_{\text{global}} = \frac{1}{L}\sum_{i=1}^{L} \mathbf{H}_{\text{MLLM}}[:, i, :] \in \mathbb{R}^{B \times 1 \times D}.
\end{equation}
This global context is then broadcast to match the number of geometric tokens: $\mathbf{S}_{\text{broadcast}} = \text{expand}(\mathbf{s}_{\text{global}}, N_{\text{patches}}) \in \mathbb{R}^{B \times N_{\text{patches}} \times D}$. For each geometric token position $j$, we compute a position-specific gating weight by concatenating the semantic context with the geometric feature and passing through a gating network:
\begin{equation}
\mathbf{g}_j = \sigma\!\left(\mathbf{W}_{\text{gate}}[\mathbf{S}_{\text{broadcast}}[:, j, :];\, \mathbf{F}_{\text{geo}}[:, j, :]]\right) \in \mathbb{R}^{B \times D},
\end{equation}
where $\sigma$ denotes the sigmoid activation, $[\cdot;\cdot]$ denotes concatenation, and $\mathbf{W}_{\text{gate}} \in \mathbb{R}^{D \times 2D}$ is a learnable linear layer.

\textbf{Adaptive Feature Fusion.}
The fused representation at each position is obtained through weighted blending of independently projected semantic and geometric features:
\begin{equation}
\mathbf{f}_{\text{fused}, j} = \mathbf{g}_j \odot \mathbf{W}_s \mathbf{S}_{\text{broadcast}}[:, j, :] + (1-\mathbf{g}_j) \odot \mathbf{W}_g \mathbf{F}_{\text{geo}}[:, j, :],
\end{equation}
where $\mathbf{W}_s, \mathbf{W}_g \in \mathbb{R}^{D \times D}$ are learnable projection matrices, and $\odot$ denotes element-wise multiplication.

The complete set of fused geometric tokens $\mathbf{F}_{\text{fused}} = \{\mathbf{f}_{\text{fused}, j}\}_{j=1}^{N_{\text{patches}}} \in \mathbb{R}^{B \times N_{\text{patches}} \times D}$ is then concatenated with the original MLLM hidden states to form an enriched conditioning sequence:
\begin{equation}
\mathbf{H}_{\text{cond}} = [\mathbf{H}_{\text{MLLM}};\, \mathbf{F}_{\text{fused}}] \in \mathbb{R}^{B \times (L + N_{\text{patches}}) \times D}.
\end{equation}

\begin{figure}[h]
    \centering
    \includegraphics[width=1.0\textwidth]{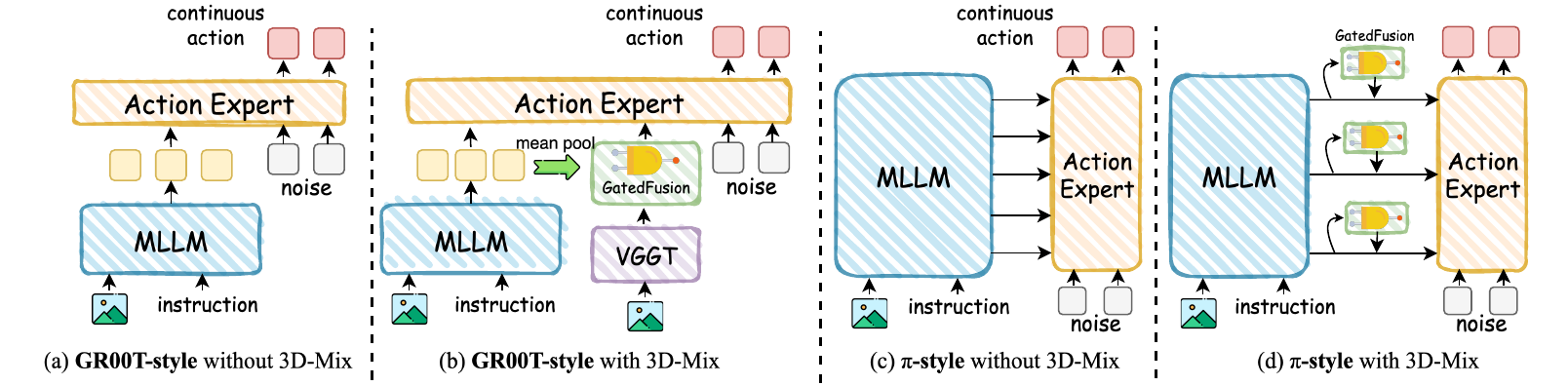}
    \caption{\textbf{Integration of \methodname{} into GR00T-style and $\pi$-style VLA Architectures}}
    \label{fig:vla-arch}
\end{figure}

\subsection{Integration with GR00T-style Architecture}

GR00T-style VLA models employ a modular architecture that separates semantic reasoning from action generation (detailed in Appendix~\ref{sec:appendix_groot}). The MLLM backbone processes multi-view observations and language instructions to extract semantic features, which are then consumed by a Diffusion Transformer (DiT)-based flow-matching action expert for continuous action prediction.

For this architecture, \methodname{} is inserted as a bridge module between the MLLM and the action expert. The MLLM produces final-layer hidden states $\mathbf{H}_{\text{MLLM}} \in \mathbb{R}^{B \times L \times D}$, which are passed through \methodname{} along with VGGT features to produce the enriched conditioning sequence $\mathbf{H}_{\text{cond}} = [\mathbf{H}_{\text{MLLM}};\, \mathbf{F}_{\text{fused}}] \in \mathbb{R}^{B \times (L + N_{\text{patches}}) \times D}$. The DiT action expert then performs cross-attention over $\mathbf{H}_{\text{cond}}$ during denoising:
\begin{equation}
\mathbf{A}_{\tau} = \text{DiT}(\mathbf{A}_{\tau-1}, \mathbf{H}_{\text{cond}}, \tau),
\end{equation}
where $\mathbf{A}_{\tau}$ represents the action sequence at diffusion timestep $\tau$. This integration preserves the standard DiT architecture with no modifications required.

\subsection{Integration with $\pi$-style Architecture}

$\pi$-style VLA models employ a layer-wise cross-attention mechanism that enables deep feature alignment between the MLLM backbone and the action expert (detailed in Appendix~\ref{sec:appendix_pi}). Unlike GR00T-style models that use only the final MLLM layer, $\pi$-style architectures extract hidden states from multiple MLLM layers and establish cross-attention connections at each corresponding DiT layer.

For $\pi$-style architectures, \methodname{} must be adapted to support layer-wise fusion. Given MLLM hidden states from multiple layers $\{\mathbf{H}^{(1)}_{\text{MLLM}}, \mathbf{H}^{(2)}_{\text{MLLM}}, \dots, \mathbf{H}^{(N)}_{\text{MLLM}}\}$, where $N$ is the number of DiT layers, we apply \methodname{} independently at each layer. Specifically, VGGT features are extracted once and projected to $\mathbf{F}_{\text{geo}} \in \mathbb{R}^{B \times N_{\text{patches}} \times D}$, which are then reused across all layers. For each layer $i$, we compute a layer-specific semantic context:
\begin{equation}
\mathbf{s}^{(i)}_{\text{global}} = \frac{1}{L}\sum_{j=1}^{L} \mathbf{H}^{(i)}_{\text{MLLM}}[:, j, :],
\end{equation}
and apply the gating mechanism to produce layer-specific fused features:
\begin{equation}
\mathbf{F}^{(i)}_{\text{fused}} = \text{GatedFusion}(\mathbf{s}^{(i)}_{\text{global}}, \mathbf{F}_{\text{geo}}),
\end{equation}
where each layer has its own gating network parameters. The enriched conditioning sequence for layer $i$ becomes:
\begin{equation}
\mathbf{H}^{(i)}_{\text{cond}} = [\mathbf{H}^{(i)}_{\text{MLLM}};\, \mathbf{F}^{(i)}_{\text{fused}}] \in \mathbb{R}^{B \times (L + N_{\text{patches}}) \times D}.
\end{equation}

The DiT action expert then performs layer-wise cross-attention, where the $i$-th DiT transformer block attends to $\mathbf{H}^{(i)}_{\text{cond}}$:
\begin{equation}
\mathbf{Z}^{(i)} = \text{TransformerBlock}^{(i)}(\mathbf{Z}^{(i-1)}, \mathbf{H}^{(i)}_{\text{cond}}),
\end{equation}
where $\mathbf{Z}^{(i)}$ represents the action latent at layer $i$.

This layer-wise fusion strategy offers two advantages: (1) \textit{Hierarchical spatial reasoning}: different layers can focus on different levels of geometric abstraction; (2) \textit{Adaptive layer-specific fusion}: each layer learns its own gating parameters for the optimal fusion at its abstraction level.


\subsection{Implementation as a Plug-and-Play Module}

The \methodname{} module requires only two inputs: (1) MLLM hidden states $\mathbf{H}_{\text{MLLM}}$ and (2) VGGT-extracted geometric features $\mathbf{F}_{\text{VGGT}}$, with no modifications to the MLLM or action expert source code.

\textbf{Integration Protocol.} (1) Load a pre-trained VGGT model and freeze its parameters; (2) Insert \methodname{} between the MLLM and action expert, configuring for either GR00T-style (single-layer) or $\pi$-style (layer-wise) fusion; (3) Train end-to-end with standard flow-matching objectives, allowing gradients to flow through the fusion module while keeping VGGT frozen.

\section{Experiment}
\label{sec:experiment}

\subsection{Experiment Setup}
\label{subsec:experiment_setup}

We evaluate \methodname{} on both GR00T-style and $\pi$-style VLA architectures. For GR00T-style, we cover six MLLM series across nine model variants spanning 2B--8B parameters: Qwen2.5-VL, Qwen3-VL, RoboBrain2.0~\citep{RoboBrain2}, RoboBrain2.5~\citep{RoboBrain2_5}, MimoEmbodied~\citep{Mimo-Embodied_25}, and RynnBrain~\citep{RynnBrain_26}. For $\pi$-style, we evaluate a representative subset of five model variants covering the same MLLM families. We follow the same training and evaluation protocol as in the pilot study, except that the per-device batch size is reduced from 16 to 10 for $\pi$-style models due to their higher memory demands from layer-wise cross-attention. VGGT-1B~\citep{VGGT_2025} serves as the frozen 3D feature extractor throughout.

\subsection{Experiment Results}
\label{subsec:experiment_results}

As shown in Tables~\ref{tab:vla_performance_groot} and~\ref{tab:vla_performance_pi}, \methodname{} consistently improves performance across all evaluated MLLM backbones on both architectures. On the GR00T-style architecture, the most notable gains on the out-of-domain SIMPLER benchmark come from RynnBrain-8B ($+12.51\%$), RoboBrain2.0-7B ($+11.39\%$), MimoEmbodied-7B ($+10.41\%$), and Qwen3-VL-4B ($+10.42\%$); even strong baselines such as RoboBrain2.5-8B (64.58\%) continue to benefit. On the $\pi$-style architecture, \methodname{} likewise delivers consistent gains, with RoboBrain2.5-4B ($+2.60\%$), Qwen3-VL-4B ($+6.77\%$), and Qwen3-VL-2B ($+5.38\%$) showing clear improvements on SIMPLER. These results demonstrate that \methodname{} provides consistent gains across diverse backbone capacities, baseline strengths, and both evaluated VLA architectures.

\begin{table}[htbp]
\centering
\small
\caption{Performance comparison of various MLLM backbones with and without \methodname{} on GR00T-style VLA architecture across SIMPLER and LIBERO benchmarks.}
\label{tab:vla_performance_groot}
\setlength{\tabcolsep}{4pt}
\begin{tabular}{l@{\hspace{6pt}}cccccc}
\toprule
\multirow{2}{*}{\textbf{Method}} & \multicolumn{4}{c}{\textbf{SIMPLER Tasks}} & \multirow{2}{*}{\textbf{SIMPLER Avg}} & \multirow{2}{*}{\textbf{LIBERO Avg}} \\
\cmidrule(lr){2-5}
& Stack Green & Put Carrot & Put Spoon & Put Eggplant & & \\
\midrule
Qwen3-VL-4B-Instruct & 22.92 & 57.29 & 62.50 & 88.54 & 57.81\gainph{10.42} & 96.50\gainph{1.55} \\
\quad + \textbf{\methodname{}} & 30.21 & 61.46 & 86.46 & 94.79 & 68.23\gain{10.42} & 98.05\gain{1.55} \\ \midrule
Qwen3-VL-8B-Instruct & 31.50 & 41.70 & 83.30 & 77.10 & 58.40\gainph{4.87} & 97.30\gainph{0.85} \\
\quad + \textbf{\methodname{}} & 33.30 & 53.12 & 95.83 & 70.83 & 63.27\gain{4.87} & 98.15\gain{0.85} \\ \midrule
Qwen2.5-VL-3B-Instruct        & 8.33  & 36.46 & 58.33 & 77.09 & 45.05\gainph{2.87} & 97.30\gainph{0.90} \\
\quad + \textbf{\methodname{}} & 16.67 & 33.33 & 62.50 & 79.17 & 47.92\gain{2.87} & 98.20\gain{0.90} \\ \midrule
Qwen2.5-VL-7B-Instruct        & 15.62 & 43.75 & 68.75 & 53.12 & 45.31\gainph{1.83} & 98.10\gainph{0.40} \\
\quad + \textbf{\methodname{}} & 29.17 & 39.58 & 65.62 & 54.17 & 47.14\gain{1.83} & 98.50\gain{0.40} \\ \midrule
RoboBrain2.0-7B  & 2.50  & 24.70 & 30.80 & 93.30 & 37.83\gainph{11.39} & 97.65\gainph{0.75} \\
\quad + \textbf{\methodname{}} & 7.29  & 50.00 & 64.58 & 75.00 & 49.22\gain{11.39} & 98.40\gain{0.75} \\ \midrule
RynnBrain-8B         & 28.12 & 42.71 & 81.25 & 58.33 & 52.60\gainph{12.51} & 97.85\gainph{0.35} \\
\quad + \textbf{\methodname{}} & 29.17 & 55.21 & 87.50 & 88.54 & 65.11\gain{12.51} & 98.20\gain{0.35} \\ \midrule
RynnBrain-2B         & 17.71 & 48.96 & 78.12 & 47.92 & 48.18\gainph{8.07} & 98.45\gainph{0.05} \\
\quad + \textbf{\methodname{}} & 21.88 & 60.42 & 85.42 & 57.29 & 56.25\gain{8.07} & 98.50\gain{0.05} \\ \midrule
MimoEmbodied-7B      & 17.71 & 55.21 & 71.88 & 66.67 & 52.87\gainph{10.41} & 97.95\gainph{0.63} \\
\quad + \textbf{\methodname{}} & 28.12 & 54.17 & 79.17 & 91.67 & 63.28\gain{10.41} & 98.58\gain{0.63} \\ \midrule
RoboBrain2.5-8B      & 44.79 & 45.83 & 70.83 & 96.88 & 64.58\gainph{1.05} & 97.80\gainph{0.35} \\
\quad + \textbf{\methodname{}} & 37.50 & 52.08 & 75.00 & 97.92 & 65.63\gain{1.05} & 98.15\gain{0.35} \\ \midrule
RoboBrain2.5-4B      & 26.04 & 48.96 & 88.54 & 61.46 & 56.25\gainph{6.77} & 97.80\gainph{0.25} \\
\quad + \textbf{\methodname{}} & 35.42 & 62.50 & 87.50 & 66.67 & 63.02\gain{6.77} & 98.05\gain{0.25} \\
\bottomrule
\end{tabular}
\end{table}

\begin{table}[htbp]
\centering
\small
\caption{Performance comparison of various MLLM backbones with and without \methodname{} on $\pi$-style VLA architecture across SIMPLER and LIBERO benchmarks.}
\label{tab:vla_performance_pi}
\setlength{\tabcolsep}{4pt}
\begin{tabular}{l@{\hspace{6pt}}cccccc}
\toprule
\multirow{2}{*}{\textbf{Method}} & \multicolumn{4}{c}{\textbf{SIMPLER Tasks}} & \multirow{2}{*}{\textbf{SIMPLER Avg}} & \multirow{2}{*}{\textbf{LIBERO Avg}} \\
\cmidrule(lr){2-5}
& Stack Green & Put Carrot & Put Spoon & Put Eggplant & & \\
\midrule
Qwen3-VL-4B-Instruct             & 2.08 & 42.71 & 43.75 & 54.17 & 35.68\gainph{6.77} & 96.50\gainph{0.70} \\
\quad + \textbf{\methodname{}} & 3.12 & 41.67 & 66.67 & 58.33 & 42.45\gain{6.77} & 97.20\gain{0.70} \\ \midrule
Qwen3-VL-2B-Instruct             & 4.17 & 32.29 & 36.46 & 82.29 & 38.80\gainph{5.38} & 96.60\gainph{0.95} \\
\quad + \textbf{\methodname{}} & 10.42 & 36.46 & 44.44 & 85.42 & 44.18\gain{5.38} & 97.55\gain{0.95} \\ \midrule
Qwen2.5-VL-3B-Instruct           & 8.33 & 47.92 & 53.12 & 61.46 & 42.71\gainph{5.99} & 95.10\gainph{0.65} \\
\quad + \textbf{\methodname{}} & 13.54 & 50.00 & 62.50 & 68.75 & 48.70\gain{5.99} & 95.75\gain{0.65} \\ \midrule
RynnBrain-2B         & 17.71 & 42.71 & 84.38 & 57.29 & 50.52\gainph{1.57} & 92.95\gainph{1.30} \\
\quad + \textbf{\methodname{}} & 21.88 & 60.42 & 68.75 & 57.29 & 52.09\gain{1.57} & 94.25\gain{1.30} \\ \midrule
RoboBrain2.5-4B      & 23.96 & 60.42 & 85.42 & 71.88 & 60.42\gainph{2.60} & 97.05\gainph{0.75} \\
\quad + \textbf{\methodname{}} & 35.42 & 62.5 & 87.50 & 66.67 & 63.02\gain{2.60} & 97.80\gain{0.75} \\
\bottomrule
\end{tabular}
\end{table}

\subsection{Ablation Studies}
\label{sec:ablations}

We conduct extensive ablation studies to analyze key design choices in \methodname{}.

\begin{figure}[h]
    \centering
    \includegraphics[width=1.0\textwidth]{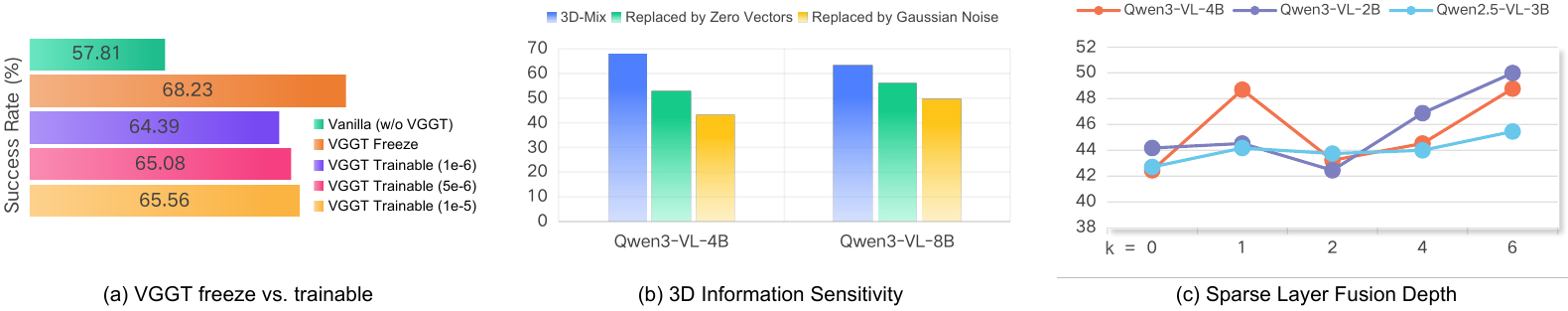}
    \caption{\textbf{Ablation study results.} Each subfigure corresponds to one ablation experiment: (a)~VGGT Frozen vs.\ Trainable. (b)~3D Information Sensitivity. (c)~Sparse Layer Fusion Depth ($\pi$-style).}
    \label{fig:ablation}
\end{figure}

\textbf{(a) VGGT Frozen vs. Trainable.} Whether to freeze or fine-tune the pre-trained VGGT model is a key design choice. As shown in Figure~\ref{fig:ablation}(a), freezing VGGT yields performance comparable to or better than fine-tuning, indicating that its pre-trained geometric representations transfer well to manipulation tasks without task-specific adaptation.

\textbf{(b) 3D Information Sensitivity.} We corrupt VGGT features at inference time by (1) replacing them with zero vectors and (2) substituting random Gaussian noise. As shown in Figure~\ref{fig:ablation}(b), both strategies cause consistent performance degradation, confirming that \methodname{}'s gains stem from genuine 3D geometric information rather than increased feature dimensionality.

\textbf{(c) Sparse Layer Fusion Depth ($\pi$-style).} We investigate sparse fusion where 3D features are injected every $k$ skipped layers ($k=0$ is full layer-wise fusion). As shown in Figure~\ref{fig:ablation}(c), sparse fusion achieves comparable or better performance than full fusion at lower memory cost, making it a practical choice for resource-constrained $\pi$-style deployments.



\section{Conclusion}
\label{sec:conclusion}

We presented \textbf{\methodname{}}, a principled plug-and-play module for integrating VGGT-based 3D geometric information into Vision-Language-Action models. A comprehensive pilot study of nine fusion schemes identified \textbf{GatedFusion} (semantic-conditioned adaptive gating) as the most effective strategy among the evaluated fusion schemes for combining 2D semantic and 3D geometric features. \methodname{} achieves consistent gains across six MLLM series on both the out-of-domain SIMPLER and in-domain LIBERO benchmarks, averaging $+7.0\%$ on SIMPLER across all nine GR00T-style variants without modifying any existing MLLM or action expert components.

Our central finding is that effective 3D integration depends not on architectural complexity, but on \emph{where} the fusion occurs and \emph{how} the balance between semantic and geometric information is set. Among the nine fusion schemes evaluated in our pilot study, semantic-conditioned gating achieved the strongest and most consistent performance gains across both in-domain and out-of-domain benchmarks, a finding further validated by its robust improvements across six diverse MLLM series. We hope \methodname{} serves as a practical and extensible foundation for future work on spatially-aware robot learning.

\bibliography{custom}

\newpage

\appendix

\section{VLA Architecture Details}
\label{sec:appendix_arch}

\subsection{GR00T-style Architecture}
\label{sec:appendix_groot}

The GR00T-style architecture adopts a modular dual-system design that decouples high-level semantic perception from low-level action generation. As illustrated in Figure~\ref{fig:vla-arch} (left), the model has two principal components: a frozen or fine-tuned MLLM backbone for visual-language encoding, and a Diffusion Transformer (DiT)-based flow-matching action expert for continuous action prediction.

\textbf{Visual-Language Encoding.}
The MLLM backbone accepts multi-view RGB images $\mathcal{I} = \{I_1, \dots, I_V\}$ and a natural language instruction $\ell$ as inputs. Images are tokenized by the vision encoder and interleaved with language tokens to form a unified input sequence. The MLLM processes this sequence through its transformer layers and produces the final-layer hidden states $\mathbf{H} \in \mathbb{R}^{B \times L \times D}$, where $B$ is the batch size, $L$ is the sequence length, and $D$ is the hidden dimension. Only the last-layer hidden states are extracted as the conditioning signal for the action expert, making this architecture compatible with any MLLM that exposes its final hidden states.

\textbf{Flow-Matching Action Expert.}
The action expert is a DiT-based network that predicts a chunk of future actions $\mathbf{A} = [\mathbf{a}_t, \dots, \mathbf{a}_{t+T}] \in \mathbb{R}^{B \times T \times d_a}$ through a flow-matching objective. At each denoising step, the DiT receives a noisy action sequence and attends to the MLLM hidden states $\mathbf{H}$ via cross-attention:
\begin{equation}
    \mathbf{Z}^{(i)} = \text{TransformerBlock}^{(i)}\!\left(\mathbf{Z}^{(i-1)},\; \mathbf{H}\right),
\end{equation}
where $\mathbf{Z}^{(i)}$ is the action latent at the $i$-th DiT layer and $\mathbf{H}$ serves as the key-value source for all layers uniformly. During training, noise $\boldsymbol{\epsilon} \sim \mathcal{N}(\mathbf{0}, \mathbf{I})$ is linearly interpolated with ground-truth actions: $\mathbf{A}_\tau = (1-\tau)\boldsymbol{\epsilon} + \tau\mathbf{A}$, where $\tau \sim \text{Beta}(\alpha, \beta)$. The model is trained to predict the velocity $\mathbf{v} = \mathbf{A} - \boldsymbol{\epsilon}$ via MSE loss. At inference, actions are recovered through Euler integration over $N$ denoising steps from random noise.

\textbf{Key Properties.}
The GR00T-style architecture is characterized by its simplicity and modularity: (1) the MLLM and action expert are loosely coupled through a single cross-attention interface, enabling independent upgrades of either component; (2) the action expert conditions on a single fixed representation $\mathbf{H}$, keeping the cross-attention cost constant regardless of MLLM depth; (3) the architecture is backbone-agnostic, so any MLLM that produces a sequence of hidden states can serve as the perception module without modification.

\subsection{$\pi$-style Architecture}
\label{sec:appendix_pi}

The $\pi$-style architecture establishes a tighter coupling between the MLLM backbone and the action expert through layer-wise cross-attention, inspired by the $\pi_0$ and $\pi_{0.5}$ model family. Rather than conditioning the action expert on a single final-layer representation, this design routes each MLLM layer's hidden states to the corresponding DiT layer, enabling hierarchical semantic-to-action alignment.

\textbf{Multi-Layer Hidden State Extraction.}
Given an MLLM with $N_{\text{vlm}}$ transformer layers, the model extracts hidden states from the last $N_{\text{dit}}$ layers (matching the number of DiT transformer blocks):
\begin{equation}
    \{{\mathbf{H}^{(1)}, \mathbf{H}^{(2)}, \dots, \mathbf{H}^{(N_{\text{dit}})}}\} = \text{MLLM}_{\text{last-}N_{\text{dit}}\text{-layers}}(\mathcal{I}, \ell),
\end{equation}
where $\mathbf{H}^{(i)} \in \mathbb{R}^{B \times L \times D}$ is the hidden state from the $i$-th selected MLLM layer. This extraction is performed in a single forward pass with \texttt{output\_hidden\_states=True}, incurring no additional computational overhead beyond storing intermediate activations.

\textbf{Layer-wise Cross-Attention.}
Each DiT transformer block attends to its corresponding MLLM layer's hidden states:
\begin{equation}
    \mathbf{Z}^{(i)} = \text{TransformerBlock}^{(i)}\!\left(\mathbf{Z}^{(i-1)},\; \mathbf{H}^{(i)}\right), \quad i = 1, \dots, N_{\text{dit}}.
\end{equation}
This layer-wise correspondence creates a hierarchical information pathway: shallow DiT layers receive lower-level MLLM representations that encode local visual features and syntactic structure, while deep DiT layers receive higher-level representations that encode semantic intent and spatial relationships. The action expert can thus leverage different levels of abstraction at different stages of the denoising process.

\textbf{Key Properties.}
Compared to the GR00T-style architecture, the $\pi$-style design offers richer semantic-to-action grounding at the cost of higher memory consumption (storing $N_{\text{dit}}$ sets of hidden states) and increased cross-attention computation. The architecture remains backbone-agnostic: the number of extracted MLLM layers is dynamically determined from the model configuration at initialization, making it compatible with MLLMs of varying depths. The layer-wise design also provides a natural integration point for \methodname{}, where 3D geometric features can be fused independently at each layer to provide spatially-enriched conditioning signals throughout the entire denoising process.

\section{Detailed VGGT Fusion Schemes}
\label{sec:appendix_fusion_details}

This section provides detailed technical descriptions of the nine VGGT fusion schemes evaluated in our pilot study (Section~\ref{sec:pilot_study}).

\textbf{(1) AE-Fusion}: Late fusion via dual cross-attention.
AE-Fusion augments the DiT-based action expert with a secondary cross-attention head that attends to projected VGGT features $\mathbf{F}'_{\text{VGGT}} = \mathbf{W}_{\text{proj}}\mathbf{F}_{\text{VGGT}}$ in parallel with the standard MLLM cross-attention, enabling simultaneous querying of semantic and geometric information during denoising. This late-fusion strategy requires architectural modifications to the action head but leaves both upstream processing streams unchanged.

\textbf{(2) Early Fusion}: Direct token-level injection at MLLM input.
EarlyFusion injects VGGT geometry tokens directly into the MLLM's input embedding sequence prior to Transformer processing. Given raw VGGT features $\mathbf{F}_{\text{VGGT}} \in \mathbb{R}^{B \times N \times D'}$, a linear projector maps them to the MLLM hidden dimension and the resulting tokens are concatenated with the standard input embeddings to form $\mathbf{X}_{\text{fused}} = [\mathbf{X}_{\text{input}};\, \mathbf{W}_{\text{proj}}\mathbf{F}_{\text{VGGT}}] \in \mathbb{R}^{B \times (L+N) \times D}$. The full MLLM Transformer then processes this combined sequence, enabling implicit semantic–geometric interaction via self-attention, with the resulting hidden states passed directly to the standard Action Expert.

\textbf{(3) Concat Fusion \& (4) CrossAttn Fusion}: GateMixer-based geometry fusion with two aggregation variants.
Both schemes share the same GateMixer preprocessing: VGGT raw tokens are passed through a patch-level module that splits the concatenated VGGT output into frame-specific and global geometric components, fuses them via learnable gating, and projects the result to obtain $\mathbf{F}_{\text{geo}} \in \mathbb{R}^{B \times N \times D}$.
The two schemes differ only in how $\mathbf{F}_{\text{geo}}$ is subsequently fused with MLLM hidden states $\mathbf{H}_{\text{MLLM}}$: \textbf{Concat Fusion} directly concatenates them—$\mathbf{H}_{\text{fused}} = [\mathbf{H}_{\text{MLLM}};\, \mathbf{F}_{\text{geo}}]$—leaving cross-modal interaction to be implicitly learned by the downstream action expert; \textbf{CrossAttn Fusion} instead applies an explicit residual cross-attention $\mathbf{F}'_{\text{geo}} = \text{CrossAttn}(\mathbf{F}_{\text{geo}}, \mathbf{H}_{\text{MLLM}}, \mathbf{H}_{\text{MLLM}}) + \mathbf{F}_{\text{geo}}$ before concatenation.

\textbf{(5) Gated Fusion}: Semantic-conditioned learnable gating between MLLM output and action expert.
GatedFusion projects VGGT tokens to the MLLM hidden dimension and derives a global semantic context by mean-pooling the MLLM hidden states $\mathbf{H}_{\text{MLLM}}$; for each geometric token position, a gate $\mathbf{g} = \sigma\!\left(\mathbf{W}_{\text{gate}}[\mathbf{s};\, \mathbf{f}_{\text{geo}}]\right)$ is computed from the concatenated semantic–geometric pair and used to produce a weighted blend $\mathbf{f}_{\text{fused}} = \mathbf{g} \odot \mathbf{W}_s\mathbf{s} + (1-\mathbf{g}) \odot \mathbf{W}_g\mathbf{f}_{\text{geo}}$. The resulting tokens are appended to the MLLM sequence and forwarded to the standard action expert.

\textbf{(6) 3D-Tokens}: Early fusion via special token alignment.
3D-Tokens appends a learnable special token $\langle|\text{vggt}|\rangle$ to every input sequence and supervises its hidden state to encode 3D geometry via an auxiliary cosine alignment loss $\mathcal{L}_{\text{align}} = 1 - \cos(\mathbf{W}_{\text{align}}\mathbf{h}_{\text{vggt}},\, \mathbf{W}_{\text{proj}}\mathbf{f}_{\text{VGGT}})$, yielding a total objective $\mathcal{L} = \mathcal{L}_{\text{action}} + \lambda_{\text{align}}\mathcal{L}_{\text{align}}$. At inference, VGGT is not required—spatial knowledge is implicitly carried by the $\langle|\text{vggt}|\rangle$ token.

\textbf{(7) Middle Layer Injection}: Adapter-style cross-attention injection at an MLLM intermediate layer.
MidLayerInjection registers a forward hook on the $k$-th MLLM decoder layer and injects projected VGGT tokens into the intermediate hidden states via a lightweight pre-norm cross-attention: $\mathbf{H}^{(k)}_{\text{out}} = \mathbf{H}^{(k)} + \alpha \cdot \text{CrossAttn}(\text{LN}(\mathbf{H}^{(k)}),\, \mathbf{F}_{\text{geo}},\, \mathbf{F}_{\text{geo}})$, where $\alpha$ is a learnable scale factor. All subsequent MLLM layers then process representations that already incorporate 3D spatial context.

\textbf{(8) Spatial Forcing}: Training-time representation alignment with zero inference overhead.
SpatialForcing supervises a designated intermediate MLLM layer by minimizing a cosine alignment loss $\mathcal{L}_{\text{align}} = -\frac{1}{N}\sum\cos\!\left(\text{BN-MLP}(\mathbf{H}^{(k)}_{\text{vis}}),\, \mathbf{F}_{\text{VGGT}} + \mathbf{E}_{\text{pos}}\right)$ between projected MLLM visual tokens and position-augmented VGGT spatial features, yielding a total training objective $\mathcal{L} = \mathcal{L}_{\text{action}} + \alpha\mathcal{L}_{\text{align}}$. At inference, both VGGT and the alignment projector are entirely discarded.

\textbf{(9) Visual Fusion}: 2D-queries-3D cross-attention fusion at the MLLM visual token level.
Visual Fusion extracts VGGT's dedicated 3D tokens $\mathbf{T}_{\text{3D}}$ and fuses them with the MLLM's 2D visual tokens via a lightweight cross-attention module where 2D tokens serve as queries and 3D tokens serve as keys and values: $\mathbf{T}'_{\text{2D}} = \text{LN}\!\left(\mathbf{T}_{\text{2D}} + \text{CrossAttn}(\mathbf{T}_{\text{2D}},\, \mathbf{T}_{\text{3D}},\, \mathbf{T}_{\text{3D}})\right)$. The spatially-enriched 2D tokens are then passed into the MLLM backbone for standard processing.

\end{document}